\documentclass[acmtog]{acmart}

\usepackage{multirow}
\usepackage[capitalize]{cleveref}

\crefname{section}{Sec.}{Secs.}
\Crefname{section}{Section}{Sections}
\crefname{table}{Tab.}{Tabs.}
\Crefname{table}{Table}{Tables}
\crefname{figure}{Fig.}{Figs.}
\Crefname{figure}{Figure}{Figures}
\crefname{equation}{Eq.}{Eqs.}
\Crefname{equation}{Equation}{Equations}

\copyrightyear{2026}
\acmYear{2026}
\setcopyright{cc}
\setcctype{by}
\acmConference[SA Conference Papers '26]{SIGGRAPH Asia 2026 Conference Papers}{December 01--04, 2026}{Kuala Lumpur, Malaysia}
\acmBooktitle{SIGGRAPH Asia 2026 Conference Papers (SA Conference Papers '26), December 01--04, 2026, Kuala Lumpur, Malaysia}
\acmDOI{10.1145/3829340.3842349}
\acmISBN{979-8-4007-2842-6/2026/12}
\begin{document}

\title{RealMe: Immersive Volumetric Video from Minimal Captures}
\author{Chenxu Zhou}
\orcid{0009-0007-0397-6737}
\email{cxzhou35@zju.edu.cn}
\affiliation{
  \institution{State Key Laboratory of CAD\&CG, Zhejiang University}
  \city{Hangzhou}
  \country{China}
}

\author{Zhiyuan Yu}
\authornote{Corresponding author.}
\orcid{0000-0001-7220-7789}
\email{zhiyuan.yu@zju.edu.cn}
\affiliation{
  \institution{State Key Laboratory of CAD\&CG, Zhejiang University}
  \city{Hangzhou}
  \country{China}
}

\author{Yudong Jin}
\orcid{0000-0001-5257-4442}
\email{krahetx@gmail.com}
\affiliation{
  \institution{State Key Laboratory of CAD\&CG, Zhejiang University}
  \city{Hangzhou}
  \country{China}
}

\author{Yinji Shentu}
\orcid{0009-0006-1887-5982}
\email{styj@zju.edu.cn}
\affiliation{
  \institution{State Key Laboratory of CAD\&CG, Zhejiang University}
  \city{Hangzhou}
  \country{China}
}

\author{Yu Zhang}
\orcid{0000-0001-7442-375X}
\email{zhangyu@4dv.ai}
\affiliation{
  \institution{4DV.ai}
  \city{Hangzhou}
  \country{China}
}

\author{Jiaming Sun}
\orcid{0000-0003-4053-8510}
\email{suenjiaming@gmail.com}
\affiliation{
  \institution{4DV.ai}
  \city{Hangzhou}
  \country{China}
}

\author{Xiaowei Zhou}
\orcid{0000-0003-1926-5597}
\email{xwzhou@zju.edu.cn}
\affiliation{
  \institution{State Key Laboratory of CAD\&CG, Zhejiang University}
  \city{Hangzhou}
  \country{China}
}

\renewcommand{\shortauthors}{Chenxu Zhou, Zhiyuan Yu, Yudong Jin, Yinji Shentu, Yu Zhang, Jiaming Sun, and Xiaowei Zhou}

\begin{abstract}
    This paper addresses the problem of creating immersive volumetric videos from sparse and low-resolution capture resources.
    Conventional dense multi-view reconstruction methods can produce high-quality volumetric videos, but their reliance on costly capture setups makes them impractical in real-world scenarios.
    Recent approaches use generative models to relax capture requirements, but still suffer from limited resolution and 4D inconsistency.
    To overcome these constraints, we propose \textbf{\textit{RealMe}}, a unified framework built around a high-resolution video diffusion model that enhances limited captures into high-resolution (2K), 4D-consistent videos for volumetric video reconstruction.
    Specifically, RealMe constructs auxiliary data from capture videos and encodes them into latents along with layout-aware positional embeddings.
    The video diffusion model adapts factorized spatio-temporal attention to the multi-view capture layout, denoises these latents across viewpoints and time, and decodes them into enhanced videos to refine the 4D representation.
    This adaptation helps the model capture temporal dynamics and multi-view relationships, leading to more 4D-consistent generation.
    To reduce the data gap between generated videos and real captures, we apply a latent frequency loss that provides frequency-domain supervision in latent space.
    Extensive experiments on the public DNA-Rendering benchmark and our self-captured \textit{RealVV} dataset demonstrate that RealMe is able to create immersive
    volumetric videos and outperforms state-of-the-art methods in visual fidelity and 4D consistency. Project page: \mbox{\href{https://zju3dv.github.io/realme}{{\footnotesize\textcolor{blue}{https://zju3dv.github.io/realme}}}}
\end{abstract}

\begin{CCSXML}
    <ccs2012>
     <concept>
      <concept_id>10010147.10010371.10010382.10010385</concept_id>
      <concept_desc>Computing methodologies~Image-based rendering</concept_desc>
      <concept_significance>500</concept_significance>
     </concept>
    </ccs2012>
\end{CCSXML}

\ccsdesc[500]{Computing methodologies~Image-based rendering}

\keywords{video generation, novel view synthesis, detail enhancement, 4D Gaussian splatting}

\begin{teaserfigure}
    \centering
    \includegraphics[width=1\textwidth]{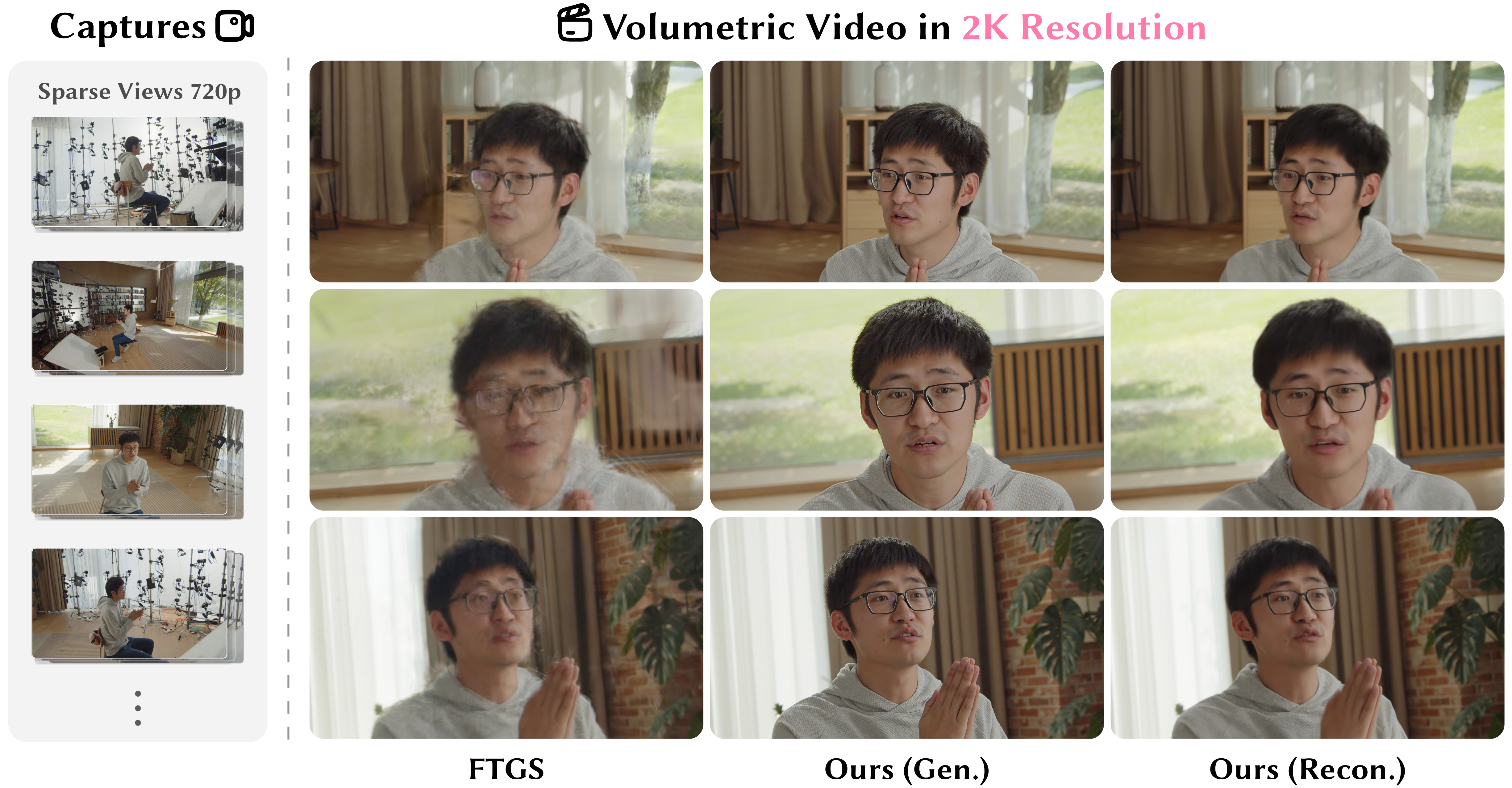}
    \caption{\textbf{RealMe creates immersive volumetric video from sparse and low-quality captures.}
    When directly optimized from sparse observations, FTGS~\cite{Wang2025FreeTimeGSFreeGaussian} struggles to maintain consistent reconstruction quality (left).
    In contrast, RealMe generates 2K multi-view videos with high visual quality and strong 4D consistency (middle), which further supports high-quality volumetric video reconstruction (right).}
    \Description{Teaser comparison. The left part shows FTGS reconstruction from sparse observations. The middle part shows RealMe generation with 2K visual quality and 4D consistency. The right part shows RealMe reconstruction as volumetric video.}
    \label{fig:teaser}
\end{teaserfigure}

\maketitle

\section{Introduction}\label{sec:introduction}

Volumetric videos capture dynamic scenes from multiple viewpoints and support free-viewpoint rendering.
The immersive and interactive nature of volumetric videos makes them an effective medium for applications such as metaverse, telepresence, VR/AR, and gaming.
Previous methods~\cite{Wu20244DGaussianSplatting,Pang2024ASHAnimatableGaussian,Xu2024RepresentingLongVolumetric,Wang2025FreeTimeGSFreeGaussian,Jiang2025TopologyAwareOptimization} have achieved high-quality volumetric video reconstruction and novel view synthesis by directly optimizing 4D scene representations.
However, these approaches typically rely on dense and well-controlled capture setups that provide sufficient multi-view observations for reconstruction.
This requirement makes them impractical under constrained acquisition settings, where only sparse camera arrays or low-quality input videos are available.
Consequently, creating volumetric videos from limited captures while approaching the visual quality and 4D consistency of dense multi-view capture remains a critical research problem and the primary motivation for our work.

The central challenge is to infer global scene structure and fine details from partial observations while preserving 4D consistency.
Recent work has explored using generative models~\cite{Brooks2024VideoGenerationModels,Esser2024ScalingRectifiedFlow,Wan2025WanOpenAdvanced} to reduce capture requirements, leveraging their learned priors for realistic, high-quality synthesis.
Some approaches~\cite{Shen2024SuperGaussian,Chen20254DSloMo4DReconstruction,Wu2025DIFIX3DImproving} adopt a two-stage optimization pipeline, using generative models to repair degraded renderings before refining the scene representation, thereby improving video resolution and frame rate.
However, their generation stage treats each viewpoint independently, ignoring multi-view correlations and causing cross-view spatial inconsistency.
The two-stage pipeline also introduces error accumulation between generation and reconstruction.
To address this limitation, another line of work~\cite{Wu2025CAT4DCreateAnything,Jin2025Diffuman4D,Huang2025Vivid4D} uses spatio-temporal attention during generation to facilitate information flow across spatial and temporal dimensions, producing more 4D-consistent results.
Despite improved consistency, these approaches still require large-scale training data and remain limited by current generative backbones,
whose native resolutions are often below 1K and therefore fall short of the visual quality required for volumetric video creation.

In this paper, we propose RealMe, a unified framework built around a high-resolution video diffusion model that enhances limited captures into high-quality, 4D-consistent videos for subsequent volumetric video reconstruction.
Specifically, given sparse and low-quality captures, RealMe first constructs auxiliary videos from input captures using diverse degradation strategies.
These auxiliary videos are encoded into conditional latents, each associated with a layout-aware positional embedding that describes its spatial, temporal, and latent-space information.
Our video diffusion model takes these conditional latents and noise as input and decodes the denoised latents into enhanced videos.
Building on prior factorized video attention~\cite{Bertasius2021TimeSformer}, we adapt the factorization to the sampled video groups, allowing denoising to propagate information separately along the temporal and spatial dimensions rather than over a flattened token sequence.
As shown in \cref{sec:ablation_attention}, this adaptation improves visual fidelity and downstream 4D consistency over full attention.
The generated videos and the original captures are used to optimize the scene representation~\cite{Wang2025FreeTimeGSFreeGaussian,Wu20244DGaussianSplatting} for volumetric video reconstruction.
To reduce the data gap between generated videos and real captures, we apply a latent frequency loss that guides the predicted latent to preserve low-frequency structure while recovering high-frequency details.

Although existing public benchmarks~\cite{Cheng2023DNARendering,Isik2023HumanRF} provide high-quality multi-view data, they are mostly captured in object-centric studio settings with the same focal-length multi-view camera arrays and do not fully cover long focal-length views or diverse real-world degradations.
To evaluate our method under more realistic and challenging acquisition settings, we curate a self-captured dataset dubbed \textit{RealVV}, which contains multi-view videos with varying focal lengths, diverse scenes, actors, and performances, as well as both real and synthetic degradations.
We evaluate RealMe on both the public DNA-Rendering benchmark~\cite{Cheng2023DNARendering} and \textit{RealVV} under limited capture settings.
Extensive experiments show that RealMe is capable of creating high-quality volumetric videos and outperforms state-of-the-art methods in visual fidelity and 4D consistency.

Our contributions are summarized as follows:
\begin{itemize}
    \item We propose RealMe, a unified framework for high-resolution 4D enhancement that reduces the capture requirements for high-quality volumetric video reconstruction.
    \item We adapt a pretrained high-resolution image diffusion model \cite{BlackForestLabs2024FLUX1Dev} to 4D enhancement using layout-aware positional embeddings and factorized spatio-temporal attention. The adapted model is trained with a latent frequency loss to reduce the data gap between generated videos and real captures.
    \item We curate \textit{RealVV}, a self-captured multi-view dataset with diverse degradations to support volumetric video evaluation under challenging settings and plan to release it to the community.
\end{itemize}

\section{Related Work}\label{sec:related_work}

\subsection[Volumetric Video Reconstruction]{Dense-View 4D Reconstruction}
Existing volumetric video reconstruction methods typically start from multi-view videos and optimize a time-varying scene representation for free-viewpoint rendering.
Recent systems~\cite{Luiten2024Dynamic3DGaussians,Wu20244DGaussianSplatting,Li2024SpacetimeGaussian,Xu20244K4D} model dynamic content with persistent primitives, 4D fields, or spatio-temporal features.
To improve temporal scale and motion coverage, specialized representations~\cite{Xu2024RepresentingLongVolumetric,Wang2025FreeTimeGSFreeGaussian,Lei2025MoScaDynamicGaussian,Jiang2025TopologyAwareOptimization} handle long sequences, complex motion, casual videos, and human-centric topology.
Other methods~\cite{Yang2024Deformable3DGaussians,Lin2024GaussianFlow4D,Wu20254DFlyFast,Wang2025ShapeMotion} reduce reconstruction cost with deformation fields, dynamic particles, or lightweight monocular pipelines.
Although these methods improve reconstruction quality and efficiency, they still rely on the captured frames as direct optimization evidence.
Their performance therefore drops when the input views are sparse or visually degraded, which motivates an explicit enhancement stage before reconstruction.

\subsection{Sparse-View 4D Reconstruction}
Sparse-view 4D reconstruction uses additional priors to reduce capture requirements.
Geometry-aware and completion methods~\cite{Mihajlovic2024SplatFields,Wang2025C4D,Wang2025MonoFusion,Zhou20264C4D,Hu2025Sparse4DGS,Su2023PointCloudCompletion} use neural fields, correspondences, monocular fusion, structured point features, flow guidance, or geometric constraints to recover structure from partial observations, including limited-view optimization and four-view capture.
A related sparse-frame formulation~\cite{Shi2026Sparse4DGS} instead addresses temporal undersampling with texture-aware regularization.
Feed-forward and generative approaches~\cite{Sabathier2025LIM,Chen20254DNeX,Kong2025GenerativeSparseView,Wu2025CAT4DCreateAnything,Zhang20244Diffusion,Liu2026Orster,Jin2025Diffuman4D,Huang2025Vivid4D} reduce per-scene optimization with learned reconstruction modules or use diffusion priors, including multi-view video models and transferred 3D/video priors, to synthesize missing observations and pseudo-view supervision.
Human-centric approaches~\cite{Xu2024RelightableandAnimatable,Zheng2024GPSGaussian,Sun2025RealtimeFree,Xiao2025RoGSplatLearningRobust} further reduce capture requirements by using body structure or learned human priors.
These approaches make sparse capture more feasible, but they mainly solve reconstruction or category-specific rendering within a fixed input setting.
They do not directly address how to turn limited captured videos into higher-fidelity inputs that preserve the capture layout for a general volumetric video pipeline.

\subsection{Generative Priors for 4D Enhancement}
Generative priors offer a complementary route for improving low-quality videos before they are used for reconstruction.
Video restoration and super-resolution methods~\cite{Xu2025VideoGigaGANTowardsDetail,Wang2025SeedVRSeedingInfinity,Wang2025SeedVR2,Zhuang2025FlashVSR,Han2025DCVSR} use generative or diffusion models to recover realistic details, stabilize temporal changes, and handle real-world degradations.
However, these models are typically designed for monocular video super-resolution, where the goal is to enhance a single video stream rather than jointly enhance multi-view captures for 4D reconstruction.
Several reconstruction pipelines use video models to increase effective frame rates~\cite{Shen2024SuperGaussian,Chen20254DSloMo4DReconstruction} or repair inferior renderings~\cite{DEGS}.
While effective, these uses of video priors remain tied to individual streams or rendered outputs rather than reconstructed representations.
Other methods use image-generation models to refine reconstructed representations~\cite{Wu2025DIFIX3DImproving} but are limited to static scenes.
As a result, they do not jointly model the camera, timestamp, and image-plane layout that determines 4D consistency across capture inputs.
Large video generation and enhancement models~\cite{Brooks2024VideoGenerationModels,He2024VEnhancer,Wan2025WanOpenAdvanced} offer a strong prior for spatio-temporal synthesis, but they require layout-aware adaptation before they can be used for multi-view capture enhancement.
RealMe builds on this direction by conditioning video diffusion on the 4D capture layout.

\begin{figure*}[t]
    \centering
    \includegraphics[width=1.0\linewidth]{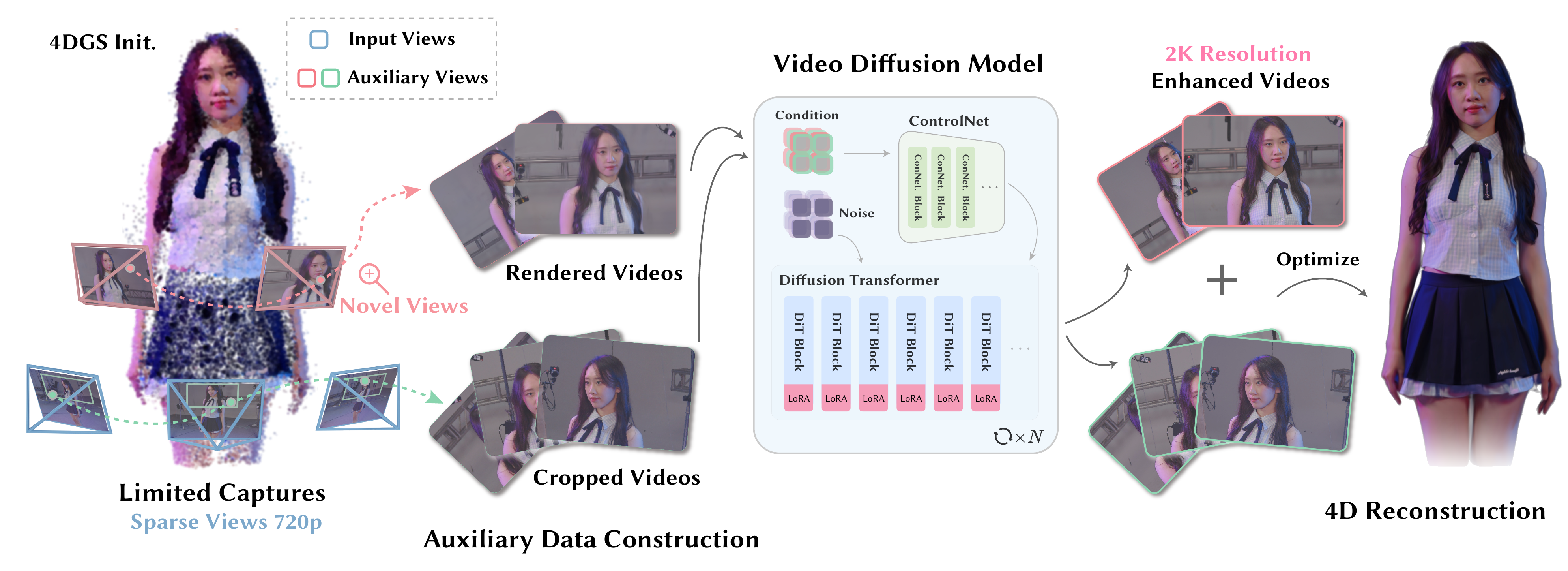}
    \caption{\textbf{Overview of our framework.}
        Given sparse and low-quality captures, RealMe first initializes a coarse 4DGS~\cite{Wang2025FreeTimeGSFreeGaussian} representation and uses it to render novel-view videos; in parallel, we detect ROIs with an offline detector~\cite{insightface} and crop the input captures to preserve local detail cues.
        These auxiliary videos are fed into a high-resolution video diffusion model to produce enhanced videos, which are then used together with the input captures to optimize the 4DGS representation for volumetric video reconstruction.
        The detailed model architecture is illustrated in \cref{fig:model_arch}.
        }\label{fig:framework}
    \Description{Overview of the RealMe framework, from limited input captures through auxiliary data construction and video enhancement to 4D reconstruction.}
\end{figure*}

\section{Method}\label{sec:method}

\subsection{Overview}\label{sec:method_overview}

\begin{figure}[t]
    \centering
    \includegraphics[width=\linewidth]{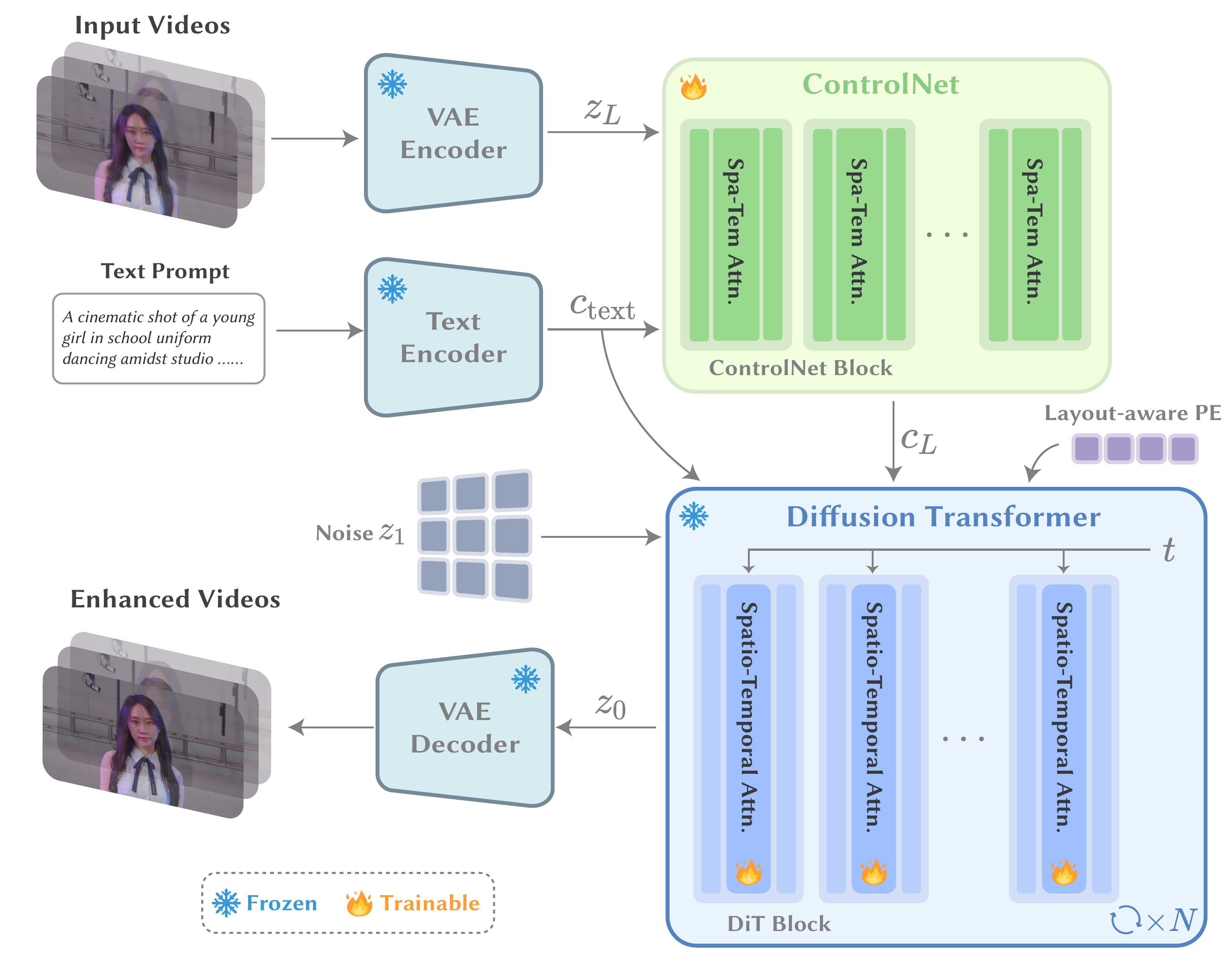}
    \caption{\textbf{Model architecture.}
        We use a ControlNet~\cite{controlnet} branch to inject control signals from condition latents and text features.
        DiT blocks use these signals with layout-aware positional embeddings to denoise noisy latents via factorized spatio-temporal attention.}
    \Description{A diagram of the video diffusion model architecture.}\label{fig:model_arch}
\end{figure}

Given sparse and low-quality captures of a dynamic scene, our target is to reconstruct a high-quality and 4D-consistent volumetric video that supports free-viewpoint rendering.
To achieve this goal, we propose RealMe, a unified 4D enhancement framework.
As illustrated in \cref{fig:framework}, RealMe first derives auxiliary videos from the input captures (\cref{sec:method_auxiliary_data}).
We then adapt the pretrained high-resolution image diffusion model~\cite{BlackForestLabs2024FLUX1Dev} to use these auxiliary videos as conditions for generating enhanced videos for 4D reconstruction (\cref{sec:method_video_diffusion}).
The detailed model architecture, including layout-aware positional embeddings and factorized spatio-temporal attention, is shown in \cref{fig:model_arch}.
To train the model, we construct paired training examples from our self-captured dataset with diverse degradations and formulate the training objectives (\cref{sec:method_training}).

\subsection{Auxiliary Data Construction}\label{sec:method_auxiliary_data}

Although the input captures are sparse and low-quality, they still contain valuable information about scene structure and motion dynamics.
We therefore convert these partial observations into auxiliary videos that guide the video diffusion model at inference time.
Let $\mathcal{V}=\{V_i\}_{i=1}^{M}$ denote the input videos and $\mathcal{P}=\{P_i\}_{i=1}^{M}$ denote their corresponding camera poses.
We first reconstruct a coarse 4DGS $\mathcal{G}_c$~\cite{Wu20244DGaussianSplatting,Wang2025FreeTimeGSFreeGaussian} from $(\mathcal{V},\mathcal{P})$.
Next, we interpolate the input camera poses into a camera trajectory to obtain $N_v$ novel-view camera poses:
\begin{equation}
    \mathcal{P}_{\mathrm{nv}}=\{P_{\mathrm{nv}}^{j}\}_{j=0}^{N_v-1}, \quad P_{\mathrm{nv}}^{j}=\mathrm{Interp}(\mathcal{P}, j/(N_v-1)).
\end{equation}
Here, the subscript $\mathrm{nv}$ denotes a novel view, and $\mathrm{Interp}(\cdot)$ denotes a pose-sampling function along the fitted camera trajectory.
Using the sampled camera poses $\mathcal{P}_{\mathrm{nv}}$, we render novel-view videos from the coarse 4DGS $\mathcal{G}_c$:
\begin{equation}
    \mathcal{V}_{\mathrm{nv}}=\{V_{\mathrm{nv}}^{j}\}_{j=0}^{N_v-1}=\mathrm{Render}(\mathcal{G}_c,\mathcal{P}_{\mathrm{nv}}).
\end{equation}
Since these videos are rendered from a coarse yet 4D-consistent representation, they provide approximate multi-view geometry guidance.
For the original input videos, we also detect regions of interest (ROIs) with an offline detector~\cite{insightface} and crop them to obtain a second set of auxiliary videos:
\begin{equation}
    \mathcal{B}=\mathrm{Det}(\mathcal{V}), \quad \mathcal{V}_{\mathrm{crop}}=\mathrm{Crop}(\mathcal{V},\mathcal{B}).
\end{equation}
Here, $\mathcal{B}$ denotes the 2D ROI boxes.
These ROI videos extract local detail cues from the input captures, while their details remain limited by the camera resolution.
We organize $\mathcal{V}_{\mathrm{nv}}$ and $\mathcal{V}_{\mathrm{crop}}$ as two separate condition types rather than concatenating content with different spatial coverage.
The two types are enhanced independently and converted into clean videos at a unified output resolution for subsequent 4D reconstruction.

\subsection{4D Enhancement Video Diffusion Model}\label{sec:method_video_diffusion}

Mainstream video diffusion models~\cite{Wan2025WanOpenAdvanced,CogVideoX} provide strong video priors, but their native resolutions are typically lower than the 2K resolution required by our volumetric video reconstruction pipeline.
We therefore adopt the pretrained high-resolution image diffusion model FLUX~\cite{BlackForestLabs2024FLUX1Dev} as our base model.

However, image diffusion models cannot be directly applied to our 4D enhancement task.
Since these models are originally designed for single-image generation, frame-wise or view-wise enhancement would make the denoising processes of different samples independent, ignoring temporal and multi-view dependencies.
Such independent processing may preserve local details within each frame, but it introduces appearance drift across time and geometry mismatch across viewpoints.
These inconsistencies degrade the enhanced videos and make the downstream volumetric video reconstruction less reliable.

To address this issue, we adapt the image diffusion model into a spatio-temporal video diffusion model for 4D enhancement.
We first describe the model architecture, where the auxiliary videos constructed in \cref{sec:method_auxiliary_data} are used as conditions to guide generation.
We then extend the original positional IDs with temporal and spatial indices to represent the structure of each sampled video group.
Finally, we adapt factorized spatio-temporal attention to these groups so information propagates separately along the temporal and spatial dimensions during denoising.

\begin{figure}[t]
    \centering
    \includegraphics[width=\linewidth]{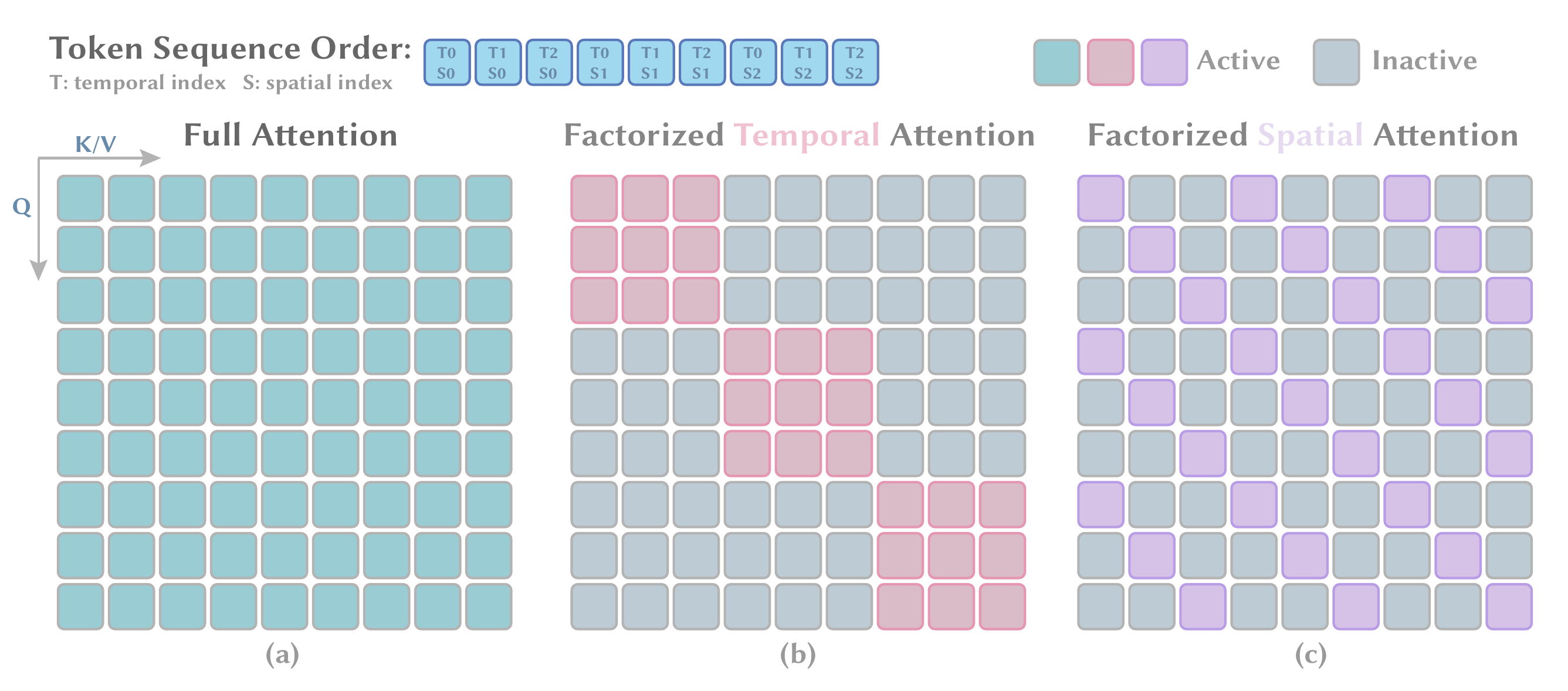}
    \caption{\textbf{Factorized spatio-temporal attention.}
        Full attention activates all sample pairs in the attention pattern, while the factorized variant keeps only pairs with the same temporal or spatial index. The horizontal and vertical axes denote key/value samples and query samples, respectively.
        Colored blocks indicate active attention regions, and gray blocks indicate inactive regions.}
    \Description{Comparison of full and factorized attention patterns over temporal and spatial sample indices.}\label{fig:facto_attn}
\end{figure}

\subsubsection{Model Architecture}

The model architecture of our video diffusion model is shown in \cref{fig:model_arch}.
Given auxiliary input videos $\mathcal{V}_{\mathrm{aux}}$ and a text prompt, we use a frozen VAE encoder to obtain condition latents $z_L$ and a frozen text encoder to obtain text features $c_{\mathrm{text}}$.
Both $z_L$ and $c_{\mathrm{text}}$ are fed into the ControlNet~\cite{controlnet} branch, which outputs control signals $c_L$ to each DiT block.
In parallel, we sample Gaussian noise $z_1$ as the initial video latent and feed it to the DiT together with the text features.
At each denoising step, the noisy latents pass through DiT blocks, where control signals are added as conditioning residuals and layout-aware positional embeddings are applied in the attention layers.
The DiT blocks then denoise the latents via factorized spatio-temporal attention, transforming $z_1$ into $z_0$ after $N$ steps.
Finally, the VAE decoder decodes $z_0$ into enhanced videos.

During training, we update only the ControlNet module and the LoRA adapters~\cite{lora} in selected DiT transformer modules.
The VAE, text encoder, and remaining pretrained DiT parameters are frozen.

\subsubsection{Layout-Aware Positional Embedding}

The original image identifiers (IDs) specify a three-axis layout coordinate, where the first axis is fixed for image tokens and the other two axes indicate the latent-grid position.
Such IDs are sufficient for single images, but they become ambiguous for sampled video groups.
Tokens from different frames or viewpoints may share the same latent-grid coordinate and therefore receive the same positional signal.

To remove this ambiguity, we extend the original three-axis IDs to four-dimensional layout-aware IDs.
For a latent token at frame $t$, view $s$, and grid coordinate $(u,v)$, we define:
\begin{equation}
    \mathbf{id}_{t,s,u,v}=(t,s,u,v),
    \quad
    0\leq t<T,\quad 0\leq s<S.
\end{equation}
Here, $T$ and $S$ denote the number of frames and viewpoints in the sampled video group.
The first two axes represent the temporal and spatial layout, while the last two axes preserve the latent-grid position.

The positional embedding is then computed by applying rotary positional encoding~\cite{Su2024RoFormer} over the four axes:
\begin{equation}
    \mathrm{PE}(\mathbf{id}_{t,s,u,v})
    =
    \phi_t(t)\Vert\phi_s(s)\Vert\phi_u(u)\Vert\phi_v(v),
\end{equation}
where $\phi_t,\phi_s,\phi_u,\phi_v$ denote RoPE mappings for each axis and their output dimensions sum to the per-head rotary dimension.
Thus, tokens with the same latent-grid position are still distinguished by their spatial and temporal indices.

\subsubsection{Factorized Spatio-Temporal Attention}

Prior work has factorized video attention along spatial and temporal axes~\cite{Bertasius2021TimeSformer}. We apply this principle to the sampled video groups, where the spatial axis indexes viewpoints rather than image patches.
The original attention mechanism concatenates the latent tokens of all samples into a long token sequence and applies full attention over the sequence.
Although this allows information exchange across samples, the samples are originally drawn from a temporal and spatial layout, and full attention does not utilize these layout relationships.
Therefore, we replace full attention with factorized spatio-temporal attention, which decomposes information propagation into temporal and spatial attention patterns.

As shown in \cref{fig:facto_attn}, each blue square in the top row denotes the latent token sequence of each sample.
The labels $T$ and $S$ denote the temporal and spatial indices of the sample in the sampled group, respectively.
After projection, these latent tokens form query, key, and value tokens for attention.
In each attention map, the horizontal axis denotes key/value tokens from the samples and the vertical axis denotes query tokens from the samples.
Each small block therefore corresponds to a sample pair, where an active block indicates that the sample pair participates in dot-product attention and an inactive block indicates that it is excluded under the current attention pattern.
We implement this adaptation by reshaping the latent tokens before attention, so attention is performed separately along the temporal and spatial dimensions and then restored to the original shape after attention.
Our ablation study further indicates that this factorized adaptation improves the 4D consistency of generated videos over full attention.

\subsection{Model Training and Optimization}\label{sec:method_training}

\subsubsection{Training Data Curation}

To train the video diffusion model, we curate \textit{RealVV}, a high-resolution multi-view video dataset with diverse scenes, actors, performances, and camera settings, including both short-focus and long-focus views.
The training split contains 100 sequences from 20 actors.
Each sequence has 48 to 80 views, with 300 frames per view.
In total, the training split contains over 1.4 million multi-view images.
We hold out the evaluation sequences and exclude them from training.
From the remaining training sequences, we construct paired training examples.
Each pair contains a clean target video and a degraded condition video.
The target provides high-resolution supervision with consistent observations, while the degraded video follows the form of the auxiliary videos used at inference time.

We synthesize the degraded condition videos using three degradation strategies:
(1) Sparse-view and underfit reconstruction degradation.
We reconstruct a coarse 4DGS from sparse views or an intentionally underfit optimization and render novel-view videos with artifacts.
(2) Resolution degradation.
We apply cropping, resizing, and downsampling to high-resolution videos to obtain low-resolution or ROI-limited videos.
(3) Appearance and motion degradation.
Following the synthetic degradation strategy of Real-ESRGAN~\cite{Wang2021RealESRGAN}, we add random noise, compression artifacts, defocus blur, and motion blur to simulate common appearance and motion degradations in real captures.
Combined with the first two strategies, these perturbations approximate the artifacts present in auxiliary videos and real captures.

\subsubsection{Training Objectives}

We train the video diffusion model with a flow matching objective~\cite{Lipman2023FlowMatching}.
The training objective also includes a latent frequency loss.
Given a clean target latent $z_0$, a Gaussian noise latent $z_1$, and a sampled time $\tau\in[0,1]$, we construct the interpolated latent:
\begin{equation}
    z_{\tau}=(1-\tau)z_0+\tau z_1.
\end{equation}
We then supervise the model to predict the corresponding velocity:
\begin{equation}
    \mathcal{L}_{\mathrm{fm}}
    =
    \left\lVert
    v_{\theta}(z_{\tau},\tau,z_L,c_{\mathrm{text}}) - (z_1-z_0)
    \right\rVert_2^2.
\end{equation}

Specifically, the latent frequency loss provides frequency-domain supervision in latent space to reduce the data gap between generated videos and real captures.
This loss encourages the predicted latent to preserve low-frequency structure while recovering high-frequency details.
We obtain the predicted clean latent:
\begin{equation}
    z_{\mathrm{pred}} = z_{\tau} - \tau v_{\theta}(z_{\tau},\tau,z_L,c_{\mathrm{text}}).
\end{equation}
Let $z_0$ denote the ground-truth clean latent, and let $\Phi_{\mathrm{L}}(\cdot)$ and $\Phi_{\mathrm{H}}(\cdot)$ denote the timestep-weighted low- and high-frequency magnitude components of a latent FFT, respectively.
The latent frequency loss is defined as:
\begin{equation}
    \mathcal{L}_{\mathrm{freq}}
    =
    \left\lVert \Phi_{\mathrm{L}}(z_{\mathrm{pred}}) - \Phi_{\mathrm{L}}(z_0) \right\rVert_1
    +
    \left\lVert \Phi_{\mathrm{H}}(z_{\mathrm{pred}}) - \Phi_{\mathrm{H}}(z_0) \right\rVert_1.
\end{equation}
The detailed construction of these frequency components and their timestep-dependent weights is provided in the supplementary material.
The final training objective is given by:
\begin{equation}
    \mathcal{L}
    =
    \mathcal{L}_{\mathrm{fm}}
    +
    \lambda_{\mathrm{freq}}\mathcal{L}_{\mathrm{freq}},
\end{equation}
where $\lambda_{\mathrm{freq}}$ is the frequency loss weight that controls the strength of frequency-domain supervision during training.

\subsubsection{4D Reconstruction}

The enhanced videos generated by the video diffusion model are used together with the input captures to refine the scene representation and reconstruct the volumetric video.
For both scene initialization and reconstruction, we adopt FTGS~\cite{Wang2025FreeTimeGSFreeGaussian} as the underlying 4DGS method, which supports dynamic scene reconstruction with time-varying Gaussian primitives.
The final 4DGS is jointly supervised by the original captures, enhanced novel-view videos, and enhanced ROI videos.
We map each enhanced ROI video back to its full-frame camera coordinates using the recorded crop boxes and optimize a unified 4D Gaussian representation for the full scene.

\section{Implementation Details}\label{sec:implementation}

We use the DiffSynth-Studio~\cite{DiffSynthStudio} framework for model training and inference.
The pretrained base model is the 14B FLUX.1 [dev] model~\cite{BlackForestLabs2024FLUX1Dev}.
We initialize the ControlNet~\cite{controlnet} branch from the pretrained DiT weights and train the full branch during adaptation.
During training, we use mixed temporal, spatial, and joint sampling, with detailed model and sampling configurations provided in the supplementary material.
We train the model for 96 hours on 8 NVIDIA H100 GPUs with an effective batch size of 8 and a learning rate of $10^{-4}$, and set the frequency loss weight to 0.1.
To reduce memory usage, we use bfloat16 mixed-precision training and gradient checkpointing.
Training target videos are generated at $2560\times1440$ resolution, while condition videos are resized to $1280\times720$ resolution.

For volumetric video reconstruction, we implement the 4DGS optimization based on EasyVolcap~\cite{Xu2023EasyVolcap}.
The training settings follow FTGS~\cite{Wang2025FreeTimeGSFreeGaussian}.
For each sequence, the final 4DGS is supervised by eight original capture views and 40 generated views enhanced from the novel-view and ROI conditions.

\section{Experiments}

\subsection{Settings}

\subsubsection{Datasets}\label{sec:datasets}

We evaluate RealMe on our self-captured \textit{RealVV} dataset and the public DNA-Rendering benchmark~\cite{Cheng2023DNARendering}.
\textit{RealVV} is a high-resolution multi-view video dataset containing diverse scenes, actors, performances, and camera settings, including both short-focus and long-focus cameras.
We use seven sequences for evaluation, all excluded from the training data.
Each evaluation sequence contains 48 views, with 180 frames per view, and is evaluated at a resolution of $2560\times1440$.
We select eight spatially distributed views as input captures and use the remaining views for novel-view evaluation.
A detailed analysis of input-view settings is provided in \cref{sec:ablation_input_views}.
DNA-Rendering~\cite{Cheng2023DNARendering} contains diverse human performances with different outfits and dynamic motions.
We select four sequences for zero-shot evaluation, each with 120 frames per view, and evaluate them at $1024\times1024$ resolution.
We follow the same input and evaluation protocol as \textit{RealVV}, using eight input views and the remaining views for novel-view evaluation.

\begin{table*}[!t]
    \centering
    \caption{\textbf{Quantitative comparison.}
    We compare RealMe with optimization-based and generation-based baselines on the \textit{RealVV} and DNA-Rendering~\cite{Cheng2023DNARendering} datasets.
    RealMe outperforms the baselines in visual quality across both datasets.
    Best and second-best results are shown in \textbf{bold} and \underline{underlined}, respectively.}
    \label{tab:comparison}
    \small
    \setlength{\tabcolsep}{3.0pt}
    \begin{tabular}{llcccccc}
        \toprule
        \multirow{2}{*}{\textbf{Method}} & \multirow{2}{*}{\textbf{Type}} & \multicolumn{3}{c}{\textbf{RealVV}} & \multicolumn{3}{c}{\textbf{DNA-Rendering}~\cite{Cheng2023DNARendering}} \\
        \cmidrule(lr){3-5} \cmidrule(lr){6-8}
        & & PSNR$\uparrow$ & SSIM$\uparrow$ & LPIPS$\downarrow$ & PSNR$\uparrow$ & SSIM$\uparrow$ & LPIPS$\downarrow$ \\
        \midrule
        FTGS~\cite{Wang2025FreeTimeGSFreeGaussian}       & Optimization & 19.0491 & 0.8142 & 0.2492 & 20.6439 & \underline{0.7986} & 0.2158 \\
        DIFIX3D+~\cite{Wu2025DIFIX3DImproving}           & Generation   & 22.6373 & 0.8267 & 0.2890 & 21.9532 & 0.7920 & 0.1666 \\
        Diffuman4D~\cite{Jin2025Diffuman4D}              & Generation   & \underline{24.4387} & \underline{0.9113} & \underline{0.2105} & \underline{23.2704} & 0.7822 & \underline{0.1427} \\
        Ours                                             & Generation   & \textbf{27.2706} & \textbf{0.9393} & \textbf{0.1621} & \textbf{25.5956} & \textbf{0.8979} & \textbf{0.1261} \\
        \bottomrule
    \end{tabular}
\end{table*}

\subsubsection{Baselines}\label{sec:baselines}

We evaluate RealMe on multiple tasks to assess its performance across several aspects.
(1) Novel view synthesis measures the visual quality of rendered novel views.
We use FTGS~\cite{Wang2025FreeTimeGSFreeGaussian} as a strong 4DGS-based reconstruction baseline and DIFIX3D+~\cite{Wu2025DIFIX3DImproving} as a generative baseline for novel-view enhancement.
We additionally include 4C4D~\cite{Zhou20264C4D}, which is designed for four-camera sparse-view reconstruction, and 4K4D~\cite{Xu20244K4D}, a high-fidelity 4D view synthesis method.
(2) 4D-consistent reconstruction tests whether the enhanced videos preserve 4D consistency for high-quality 4DGS reconstruction.
We compare with Diffuman4D~\cite{Jin2025Diffuman4D}, a generation-based sparse-view reconstruction method.
(3) Detail enhancement focuses on recovering fine appearance details from low-quality captures.
We benchmark against STAR~\cite{Xie2025STAR} and FlashVSR~\cite{Zhuang2025FlashVSR}, two video super-resolution baselines for degraded video inputs.
The implementation details and settings for the baselines are described in the supplementary material.

\subsubsection{Metrics}\label{sec:metrics}

We report PSNR, SSIM, and LPIPS as quantitative metrics.
PSNR and SSIM~\cite{Wang2004ImageQualityAssessment} measure pixel-level and structural fidelity, while LPIPS~\cite{Zhang2018UnreasonableEffectiveness} measures perceptual similarity.

For 4D-consistent reconstruction, we use additional consistency metrics.
Specifically, we first optimize a 4DGS representation~\cite{Wang2025FreeTimeGSFreeGaussian} using the enhanced videos together with the input captures.
We then render videos from the camera poses corresponding to the enhanced videos and compute PSNR, SSIM, and LPIPS between the enhanced videos and the corresponding renderings.
We use the subscript $c$ to denote these metrics, e.g., PSNR$_c$.
For temporal consistency, we report the optical-flow-based Temporal Warping Error (TWE)~\cite{Lai2018LearningBlindVideoTemporalConsistency}, which measures the residual between adjacent frames after flow-based warping over non-occluded regions; lower values indicate greater temporal stability.
Because temporal smoothing can also reduce this error, we use TWE as a complementary stability metric rather than a measure of reconstruction fidelity.

\subsection{Comparison Experiments}\label{sec:comparison}

\begin{figure*}[p]
    \centering
    \includegraphics[width=\textwidth,height=0.92\textheight,keepaspectratio]{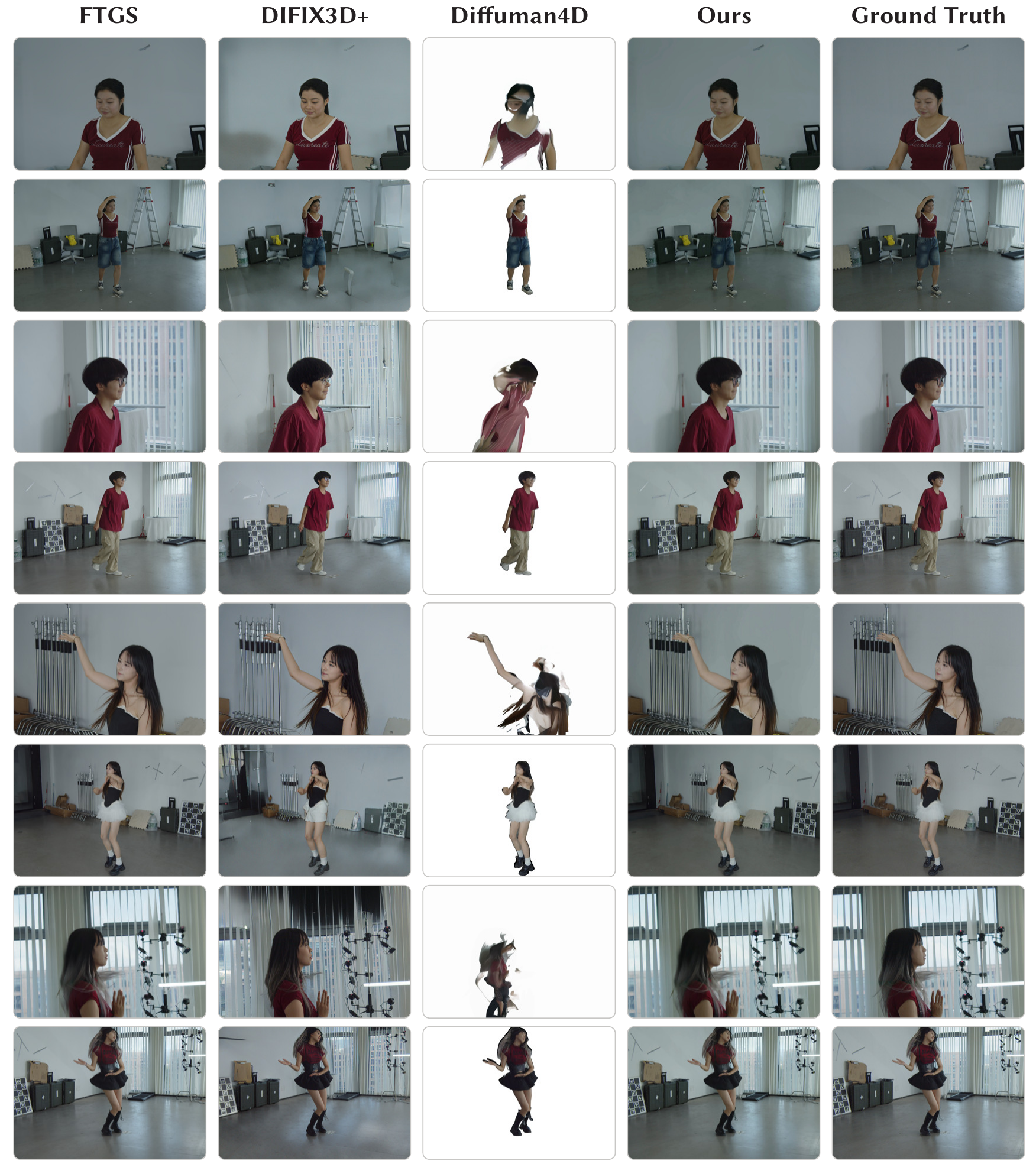}
    \caption{\textbf{Qualitative comparison for novel view synthesis on \textit{RealVV}.}
    Given the same sparse input captures, RealMe produces higher visual quality and more consistent novel views than optimization-based and generation-based baselines.
    Diffuman4D~\cite{Jin2025Diffuman4D} cannot generate scene backgrounds and generalizes poorly to zoomed-in long-focus views, while RealMe preserves both human appearance and surrounding scene content.}
    \Description{Qualitative comparison for novel view synthesis on RealVV.}
    \label{fig:nvs_realvv}
\end{figure*}

We compare RealMe with optimization-based and generation-based baselines under the same sparse-capture setting.
Quantitative results are reported in \cref{tab:comparison}, and qualitative comparisons on \textit{RealVV} are shown in \cref{fig:nvs_realvv}.
RealMe achieves the best visual quality among all compared methods across both datasets.
When optimizing the 4D representation directly from sparse input views, FTGS~\cite{Wang2025FreeTimeGSFreeGaussian} produces lower-quality novel views because the sparse observations provide insufficient constraints for unseen regions and view-dependent appearance.
In contrast, RealMe derives scene structure from sparse input views and recovers the missing appearance.
In addition to the optimization-based baseline, the generation-based baselines exhibit different limitations.
DIFIX3D+~\cite{Wu2025DIFIX3DImproving} improves local details, but it does not consistently preserve the target scene appearance.
Diffuman4D~\cite{Jin2025Diffuman4D} focuses on full-body human reconstruction and therefore cannot model scene backgrounds.
Due to the domain gap in its training data, it also does not generalize well to zoomed-in long-focus views in \textit{RealVV}.

We additionally compare with two 4D reconstruction baselines under a four-camera setting on \textit{RealVV}, as reported in \cref{tab:four_camera_comparison}.
Overall, our method demonstrates stronger reconstruction performance under the four-camera setting.
Compared with these optimization-based methods, our approach leverages generative priors to recover high-frequency details missing from sparse observations and reduce artifacts caused by sparse-view inputs.

We further evaluate the overall consistency between the generated videos and the input videos through downstream 4D reconstruction in \cref{tab:consistency_comparison}.
The results indicate that, compared with Diffuman4D~\cite{Jin2025Diffuman4D}, our generated videos better align with the input captures and lead to more coherent 4D reconstruction.
These metrics are the consistency metrics defined in \cref{sec:metrics}.
As shown in \cref{tab:twe_comparison}, RealMe achieves temporal consistency comparable to the optimization-based baseline while outperforming the generation-based methods.

\begin{figure*}[p]
    \centering
    \includegraphics[width=\textwidth,height=0.92\textheight,keepaspectratio]{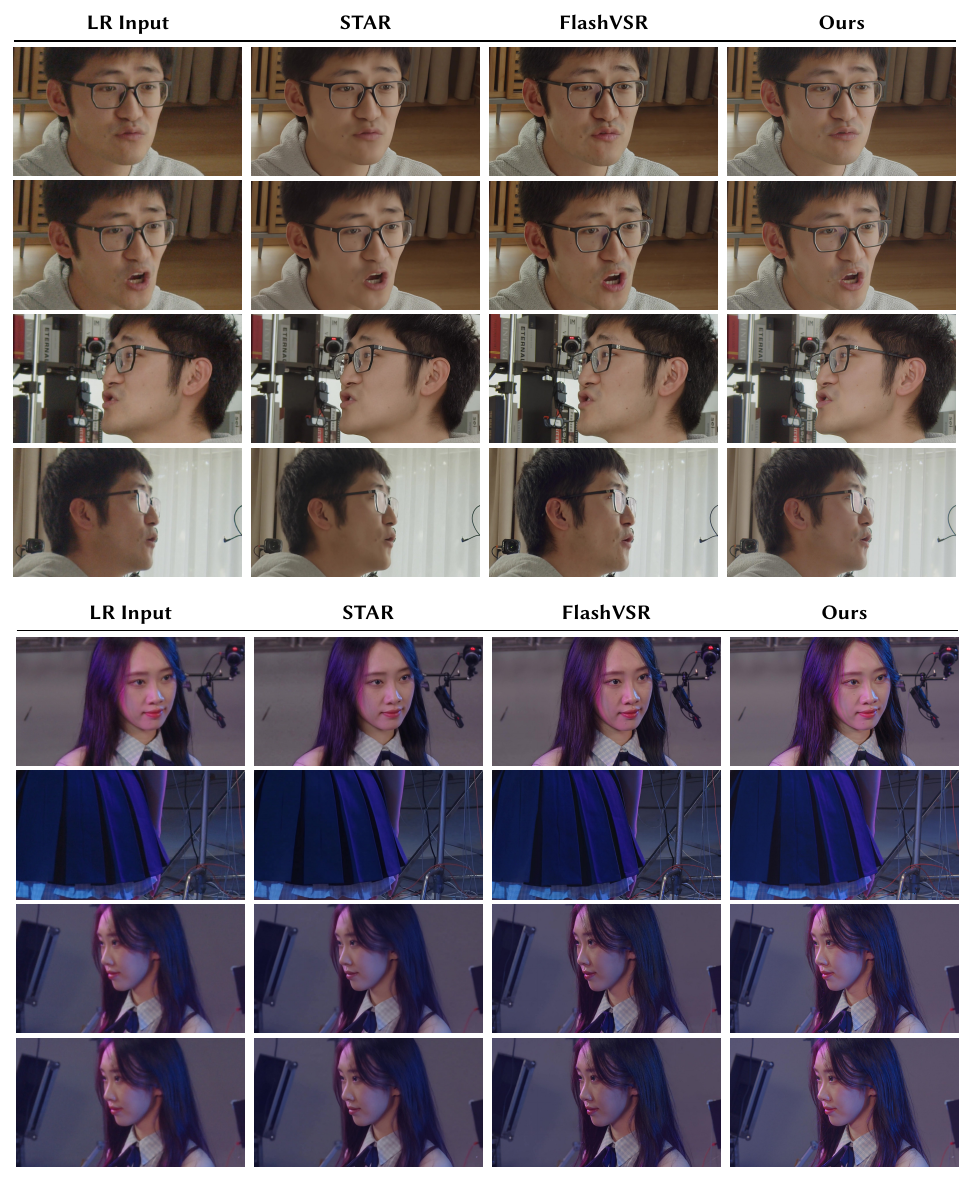}
    \caption{\textbf{Qualitative comparison for detail enhancement on \textit{RealVV}.}
    Starting from low-resolution inputs, STAR~\cite{Xie2025STAR} and FlashVSR~\cite{Zhuang2025FlashVSR} tend to produce over-smoothed regions or artificial local textures.
    In contrast, RealMe restores richer facial details, hair structures, and clothing textures with a more natural appearance.
    Please \textbf{zoom in} to better inspect the local visual quality.}
    \Description{Qualitative comparison for detail enhancement on RealVV.}
    \label{fig:detail_enhancement}
\end{figure*}

In addition, we evaluate detail enhancement at 2K resolution in \cref{fig:detail_enhancement}.
STAR~\cite{Xie2025STAR} and FlashVSR~\cite{Zhuang2025FlashVSR} improve the low-resolution inputs, but their results can still contain over-smoothed facial regions, softened hair boundaries, or artificial local textures.
In contrast, RealMe recovers finer facial, hair, and clothing details while maintaining a natural appearance.

\begin{table}[!htbp]
    \centering
    \caption{\textbf{Representative novel view synthesis results on \textit{RealVV} with four cameras.}
    All methods use four input views.
    Best results are shown in \textbf{bold}.}
    \label{tab:four_camera_comparison}
    \small
    \setlength{\tabcolsep}{4pt}
    \begin{tabular}{lccc}
        \toprule
        \textbf{Method} & PSNR$\uparrow$ & SSIM$\uparrow$ & LPIPS$\downarrow$ \\
        \midrule
        4C4D~\cite{Zhou20264C4D} & 19.0405 & 0.8682 & 0.3716 \\
        4K4D~\cite{Xu20244K4D}   & 15.6413 & 0.8181 & 0.5090 \\
        Ours                       & \textbf{24.9304} & \textbf{0.8735} & \textbf{0.2278} \\
        \bottomrule
    \end{tabular}
\end{table}

\begin{table}[!htbp]
    \centering
    \caption{\textbf{Quantitative comparison on 4D-consistent reconstruction.}
    PSNR$_c$, SSIM$_c$, and LPIPS$_c$ denote the consistency metrics defined in \cref{sec:metrics}.
    Best results are shown in \textbf{bold}.}
    \label{tab:consistency_comparison}
    \small
    \setlength{\tabcolsep}{3.0pt}
    \begin{tabular}{lccc}
        \toprule
        \textbf{Method} & PSNR$_c\uparrow$ & SSIM$_c\uparrow$ & LPIPS$_c\downarrow$ \\
        \midrule
        Diffuman4D~\cite{Jin2025Diffuman4D}              & 26.7098 & 0.8437 & 0.1913 \\
        Ours                                             & \textbf{28.0756} & \textbf{0.9246} & \textbf{0.1385} \\
        \bottomrule
    \end{tabular}
\end{table}

\begin{table}[!htbp]
    \centering
    \caption{\textbf{Temporal consistency comparison.}
    TWE is defined in \cref{sec:metrics}.
    Best and second-best results are shown in \textbf{bold} and \underline{underlined}, respectively.}
    \label{tab:twe_comparison}
    \small
    \setlength{\tabcolsep}{4pt}
    \begin{tabular}{lc}
        \toprule
        \textbf{Method} & TWE$\downarrow$ \\
        \midrule
        FTGS~\cite{Wang2025FreeTimeGSFreeGaussian} & \textbf{0.00299} \\
        DIFIX3D+~\cite{Wu2025DIFIX3DImproving}     & 0.04533 \\
        Diffuman4D~\cite{Jin2025Diffuman4D}       & 0.03170 \\
        Ours                                      & \underline{0.00551} \\
        \bottomrule
    \end{tabular}
\end{table}

\subsection{Ablation Studies}\label{sec:ablation}

We conduct ablation studies to assess the contribution of the model adaptations and training objective used in RealMe.
We additionally vary the number of input views to analyze the trade-off between capture requirements and reconstruction quality.
All ablation studies are performed on the evaluation sequences of the \textit{RealVV} dataset.

\subsubsection{Ablation on the Number of Input Views}\label{sec:ablation_input_views}

As shown in \cref{tab:ablation_input_views}, we evaluate RealMe with different numbers of input views to study the trade-off between capture requirements and reconstruction quality.
The results show that RealMe works across the evaluated input-view settings, with a substantial improvement from four to eight views and only limited additional benefits from increasing to 16 views.
We therefore use eight spatially distributed views as the default setting because they provide a practical balance between capture requirements and reconstruction quality.
In practical applications, selecting the number of input views involves the same trade-off: film-production and studio settings that prioritize reconstruction quality favor more views, whereas outdoor captures with limited equipment favor fewer views.

\begin{table}[!htbp]
    \centering
    \caption{\textbf{Ablation on the number of input views.}
    We evaluate reconstruction quality on \textit{RealVV} under different input-view settings.
    Best results are shown in \textbf{bold}.}
    \label{tab:ablation_input_views}
    \small
    \setlength{\tabcolsep}{5pt}
    \begin{tabular}{lccc}
        \toprule
        \textbf{Input Views} & PSNR$\uparrow$ & SSIM$\uparrow$ & LPIPS$\downarrow$ \\
        \midrule
        4           & 24.0490 & 0.8049          & 0.2705 \\
        8 (default) & 27.2706 & \textbf{0.9393} & 0.1621 \\
        16          & \textbf{27.3614} & 0.9159 & \textbf{0.1389} \\
        \bottomrule
    \end{tabular}
\end{table}

\subsubsection{Ablation on Model Components}\label{sec:ablation_attention}

As shown in \cref{tab:ablation_model}, we evaluate the layout-aware positional embedding and factorized spatio-temporal attention by removing each one from the full model.
The variant without layout-aware positional embedding uses the original positional embedding, while the variant without factorized spatio-temporal attention replaces it with full attention.
Removing layout-aware positional embedding lowers PSNR$_c$, suggesting that layout-aware positional signals help preserve downstream 4D consistency.
Replacing factorized spatio-temporal attention with full attention reduces both visual quality and consistency metrics, indicating that full attention over flattened tokens is less effective for the sampled video group.
Together, these ablations show that both components contribute to visual fidelity and downstream 4D consistency.

\begin{table}[!htbp]
    \centering
    \caption{\textbf{Ablation on model components.}
    LA-PE denotes layout-aware positional embedding, FSA denotes factorized spatio-temporal attention, and PSNR$_c$ denotes the consistency metric defined in \cref{sec:metrics}.}
    \label{tab:ablation_model}
    \small
    \setlength{\tabcolsep}{4pt}
    \begin{tabular}{lcccc}
        \toprule
        \textbf{Method} & PSNR$\uparrow$ & SSIM$\uparrow$ & LPIPS$\downarrow$ & PSNR$_c\uparrow$ \\
        \midrule
        w/o LA-PE & 22.7354 & 0.9095 & 0.2355 & 25.2596 \\
        w/o FSA   & 24.3178 & 0.8995 & 0.2166 & 25.8435 \\
        Ours      & \textbf{24.8924} & \textbf{0.9226} & \textbf{0.1383} & \textbf{27.2688} \\
        \bottomrule
    \end{tabular}
\end{table}

\subsubsection{Ablation on Latent Frequency Loss}\label{sec:ablation_lf_loss}

We ablate the latent frequency loss and supervision space in \cref{tab:ablation_loss} to assess the gap between generated videos and real captures.
Both image- and latent-space variants improve quality over the baseline, with the latent-space variant performing best; image-space results use 1K resolution because 2K training runs out of memory.
Additional qualitative results are provided in the supplementary material.

\begin{table}[!htbp]
    \centering
    \caption{\textbf{Ablation on latent frequency loss.}
    $\mathcal{L}_{f}$ denotes frequency loss, while $\mathcal{L}_{\mathrm{img}}$ and $\mathcal{L}_{\mathrm{lat}}$ denote image-space and latent-space frequency loss, respectively.
    LPIPS$_c$ denotes the consistency metric defined in \cref{sec:metrics}.
    Metrics for $\mathcal{L}_{\mathrm{img}}$ are computed at 1K resolution because 2K training runs out of memory.}
    \label{tab:ablation_loss}
    \small
    \setlength{\tabcolsep}{4pt}
    \begin{tabular}{lcccc}
        \toprule
        \textbf{Method} & PSNR$\uparrow$ & SSIM$\uparrow$ & LPIPS$_c\downarrow$ & GPU VRAM$\downarrow$ \\
        \midrule
        w/o $\mathcal{L}_{f}$  & 22.6017 & 0.8497 & 0.3113 & \textbf{72.4 GB} \\
        w/ $\mathcal{L}_{\mathrm{img}}$  & 24.7123 & 0.8644 & 0.2849 & \textbf{OOM (2K)} \\
        w/ $\mathcal{L}_{\mathrm{lat}}$  & \textbf{24.9641} & \textbf{0.8891} & \textbf{0.2201} & 73.5 GB \\
        \bottomrule
    \end{tabular}
\end{table}

\section{Conclusion and Discussion}\label{sec:conclusion}

We present RealMe, a unified framework that enhances sparse and low-quality input captures into high-resolution, 4D-consistent videos for volumetric video reconstruction.
The framework adapts the pretrained high-resolution image diffusion model~\cite{BlackForestLabs2024FLUX1Dev} to model temporal dependencies and spatial relationships while preserving local details.
Specifically, we construct auxiliary videos from the input captures as conditioning inputs for the video diffusion model.
It extends the original positional IDs with temporal and spatial indices and applies factorized spatio-temporal attention along these dimensions, improving the 4D consistency of generated videos.
Training further applies a latent frequency loss to reduce the data gap between generated videos and real captures.
Experiments on our self-captured dataset \textit{RealVV} and the public DNA-Rendering benchmark~\cite{Cheng2023DNARendering} show that RealMe achieves superior visual quality and 4D consistency over existing baselines, while extending volumetric video reconstruction to 2K resolution under limited capture conditions.

\paragraph{Limitations and future work.}

Although RealMe achieves high-quality volumetric video reconstruction with improved 4D consistency, it still has several limitations.
Due to the scarcity of high-quality multi-view datasets, our training data remains constrained, and the model shows limited generalization to large scenes or complex camera settings.
The method may fail when the input captures provide insufficient geometric cues to initialize a valid coarse 4DGS.
How to scale up training data and accelerate high-resolution diffusion inference remain important directions for our future work.

\begin{acks}
This work was partially supported by the National Key R\&D Program of China (No. 2024YFB2809105), the Zhejiang Provincial Natural Science Foundation of China (No. LR25F020003), and the Information Technology Center and State Key Lab of CAD\&CG, Zhejiang University.
\end{acks}

\clearpage

\bibliographystyle{ACM-Reference-Format}
\bibliography{acmart}

\clearpage
\appendix
\twocolumn[{\raggedright\Huge\sffamily Supplementary Material for RealMe: Immersive Volumetric Video from Minimal Captures\par\bigskip}]
\section{Baseline Settings}

This section details the baseline configurations used in our evaluation.
To keep the comparisons consistent, each baseline uses the same input views and frame ranges as the main experiments whenever it supports the same inputs.
We rely on official implementations and released checkpoints when available, and resize the outputs to the ground-truth resolution before computing metrics.

\paragraph{FTGS}
For the 4DGS baseline, we use FTGS~\cite{Wang2025FreeTimeGSFreeGaussian} implemented with EasyVolcap~\cite{Xu2023EasyVolcap}.
The optimization follows the training settings of the original FTGS paper.
Given the same sparse input captures as RealMe, we optimize a dynamic Gaussian representation, render the held-out views, and compute novel-view metrics against the ground-truth videos.

\paragraph{4C4D}
We use the official PyTorch implementation of 4C4D~\cite{Zhou20264C4D} and optimize a separate representation for each evaluated sequence.
We use four spatially distributed input cameras and the frame range defined by the four-camera evaluation setting.
Following the official pipeline, we initialize the Gaussian point cloud from the four training views with MASt3R~\cite{Leroy2024MASt3R} and retain the released configuration, including 30,000 optimization iterations.
We render the held-out cameras from the final checkpoint and pass the resulting full-frame images to the common novel-view evaluation pipeline.

\paragraph{4K4D}
\begingroup
\sloppy
We evaluate 4K4D~\cite{Xu20244K4D} using its official EasyVolcap-based implementation and train a separate model for each evaluated sequence.
We adapt the released custom-dataset configuration to the calibrated RGB frames and camera parameters of \textit{RealVV}, restrict the training-view list to four cameras and the corresponding evaluation frame range, and retain the official 1,600-epoch schedule and remaining model and optimization settings.
After optimization, we render the held-out camera views and evaluate the full-frame outputs with the same resizing and metric computation used for the other methods.
\par
\endgroup

\paragraph{DIFIX3D+}
We evaluate DIFIX3D+~\cite{Wu2025DIFIX3DImproving} as the diffusion-based enhancement baseline for sparse-view 3D reconstruction.
We use the official code and checkpoint, with the released \texttt{difix\_ref} model.
For each target view, the input view with the nearest camera pose is selected as the reference view.
The enhanced views are then passed to the same evaluation pipeline used by our method.

\paragraph{Diffuman4D}
We use Diffuman4D~\cite{Jin2025Diffuman4D} as the sparse-view human reconstruction baseline.
This method models sparse-view videos with spatio-temporal diffusion, and we evaluate it in a zero-shot manner using the official code and pretrained checkpoint.
Since Diffuman4D is designed for human performance reconstruction, we compute its metrics inside the foreground mask, so the evaluation focuses on the region modeled by the baseline.

\paragraph{STAR and FlashVSR}
For detail enhancement, we compare with two video super-resolution baselines, STAR~\cite{Xie2025STAR} and FlashVSR~\cite{Zhuang2025FlashVSR}.
We use their official code and released checkpoints, with input videos trimmed to the same frame range as our setting.
Because these methods generate outputs at their default resolutions, we resize the results to the ground-truth resolution before computing PSNR, SSIM, and LPIPS.

\section{Additional Implementation Details}

\subsection{Model Architecture}

Our video diffusion model is built on FLUX.1 [dev]~\cite{BlackForestLabs2024FLUX1Dev} and implemented with DiffSynth-Studio~\cite{DiffSynthStudio}.
A ControlNet branch~\cite{controlnet} injects conditioning signals from the auxiliary videos into the diffusion model.
We initialize this branch from the corresponding pretrained DiT weights, which aligns its feature scale with the base model at the start of training.
During adaptation, the full ControlNet branch is trainable.
For the pretrained DiT, we add rank-32 LoRA adapters~\cite{lora} to the QKV projections, attention output projections, MLP layers, and modulation/projection layers, while the remaining pretrained DiT weights are frozen.
In both joint and single transformer blocks, each repeated factorized-attention schedule contains three temporal attention blocks followed by one spatial attention block.
We apply temporal attention more frequently because temporal modeling is more challenging than spatial modeling in our capture setting, making the 3:1 ratio a practical trade-off.

\subsection{Training Data}

We provide additional details about how the degraded condition videos are synthesized.
(1) For sparse-view and underfit reconstruction degradation, we either uniformly sample 32, 16, or 8 views according to camera poses to reconstruct a coarse 4DGS, or train a 4DGS with all views for 3,000 iterations.
We then render the remaining views or all views as degraded conditions.
(2) For resolution degradation, we use an offline detector~\cite{insightface} to localize facial regions, pad the detected 2D boxes, and crop local ROI videos.
For blur/resize-based low-resolution generation, we apply USM sharpening and first-stage blur with kernel sizes from 7 to 21.
We sample the configured blur-kernel types with probabilities $[0.45, 0.25, 0.12, 0.03, 0.12, 0.03]$, perform stochastic resizing, and use bicubic upsampling to produce $1280\times720$ condition videos.
(3) For appearance and motion degradation, we follow the degradation design of Real-ESRGAN~\cite{Wang2021RealESRGAN} and add random noise, compression artifacts, defocus blur, and motion blur.
Together, these degradations cover geometry artifacts, resolution loss, local detail loss, appearance noise, and motion blur.

\subsection{Training Strategy}

Condition videos are fixed at $1280\times720$ resolution during training.
For target videos, we use a progressive resolution schedule that increases from $1280\times720$ to $2560\times1440$.
This schedule stabilizes high-resolution training, while the final training stage remains consistent with the $2560\times1440$ target resolution reported in the main paper.
We sample temporal, spatial, and joint training groups with probabilities $0.4$, $0.4$, and $0.2$, respectively.
Their corresponding window sizes are 4 frames, 4 views, and a $2\times2$ temporal-view group.

\subsection{Latent Frequency Loss}

We first recover the predicted clean latent $z_{\mathrm{pred}}$:
\begin{equation}
    z_{\mathrm{pred}} = z_{\tau} - \tau v_{\theta}(z_{\tau},\tau,z_L,c_{\mathrm{text}}).
\end{equation}
Let $A(z)=\lvert\operatorname{rFFT}_{2\mathrm{D}}(z)\rvert$ denote the magnitude of the 2D real FFT over the latent spatial dimensions, and let $q(z)$ be its 80th-percentile magnitude. For large tensors, we estimate this percentile from 10,000 sampled coefficients.
We define a soft low-frequency mask $M_{\mathrm{L}}(z)$ and its high-frequency complement $M_{\mathrm{H}}(z)$ as:
\begin{equation}
    M_{\mathrm{L}}(z)=\operatorname{sigmoid}\!\left(10\left(q(z)-A(z)\right)\right),
    \quad M_{\mathrm{H}}(z)=1-M_{\mathrm{L}}(z).
\end{equation}
The complementary soft masks retain frequencies around the cutoff instead of discarding a mid-frequency band. Using $\Psi_k(z)=M_k(z)\odot A(z)$ for $k\in\{\mathrm{L},\mathrm{H}\}$, we compute the unweighted branch losses:
\begin{equation}
    \mathcal{L}_{k}
    =
    \left\lVert \Psi_k(z_{\mathrm{pred}})-\Psi_k(z_0)\right\rVert_1,
    \quad k\in\{\mathrm{L},\mathrm{H}\}.
\end{equation}
The two branches are weighted according to the diffusion timestep:
\begin{equation}
    \mathcal{L}_{\mathrm{freq}}
    =
    c_t\mathcal{L}_{\mathrm{L}}+(1-c_t)\mathcal{L}_{\mathrm{H}},
    \qquad c_t=\left(\frac{t}{T}\right)^2,
\end{equation}
where $t$ and $T$ denote the current and maximum diffusion timesteps, respectively, with $t/T=\tau$.
Equivalently, we absorb these weights into the two frequency components:
\begin{equation}
    \Phi_{\mathrm{L}}(z)=c_t\Psi_{\mathrm{L}}(z),
    \qquad
    \Phi_{\mathrm{H}}(z)=(1-c_t)\Psi_{\mathrm{H}}(z).
\end{equation}
Because the weights are nonnegative, the sum of the resulting two $L_1$ terms is equivalent to the timestep-weighted branch loss above.

\subsection{4DGS Reconstruction}

Novel-view renderings and cropped ROI videos are maintained as separate condition types, enhanced independently, and converted to the same output resolution.
For each sequence, the final 4DGS is supervised by eight original capture views and 40 generated views enhanced from the novel-view and ROI conditions.
We map enhanced ROI videos back to their full-frame camera coordinates using the recorded crop boxes before reconstruction.
Raw captures and enhanced videos use the same supervision type and loss weight.
Following FTGS~\cite{Wang2025FreeTimeGSFreeGaussian}, we optimize one unified 4D Gaussian representation for the full scene and retain the remaining FTGS settings.
Foreground masks are used only for the Diffuman4D-aligned evaluation and are not used for training or the other experiments.

\subsection{Input Synchronization and Efficiency}

The method accepts unsynchronized camera videos when their time offsets are annotated. We use these offsets to associate sampled frames with a common scene timeline before constructing temporal and joint training groups.
Runtime and peak memory are measured for diffusion inference only, using one 4-frame $2560\times1440$ batch with 30 denoising steps on a single NVIDIA H100 GPU; the results are reported in \cref{tab:supp_efficiency}.

\begin{table}[t]
    \centering
    \caption{\textbf{Diffusion inference efficiency.}
    We measure one 4-frame $2560\times1440$ batch with 30 denoising steps on a single NVIDIA H100 GPU.}
    \label{tab:supp_efficiency}
    \small
    \setlength{\tabcolsep}{8pt}
    \begin{tabular}{cc}
        \toprule
        \textbf{Runtime} & \textbf{Peak Memory} \\
        \midrule
        135.1 s & 48.1 GB \\
        \bottomrule
    \end{tabular}
\end{table}

\section{Additional Qualitative Results}

\subsection{Qualitative Comparisons}

\begin{figure*}[!t]
    \centering
    \includegraphics[width=\textwidth,height=0.88\textheight,keepaspectratio]{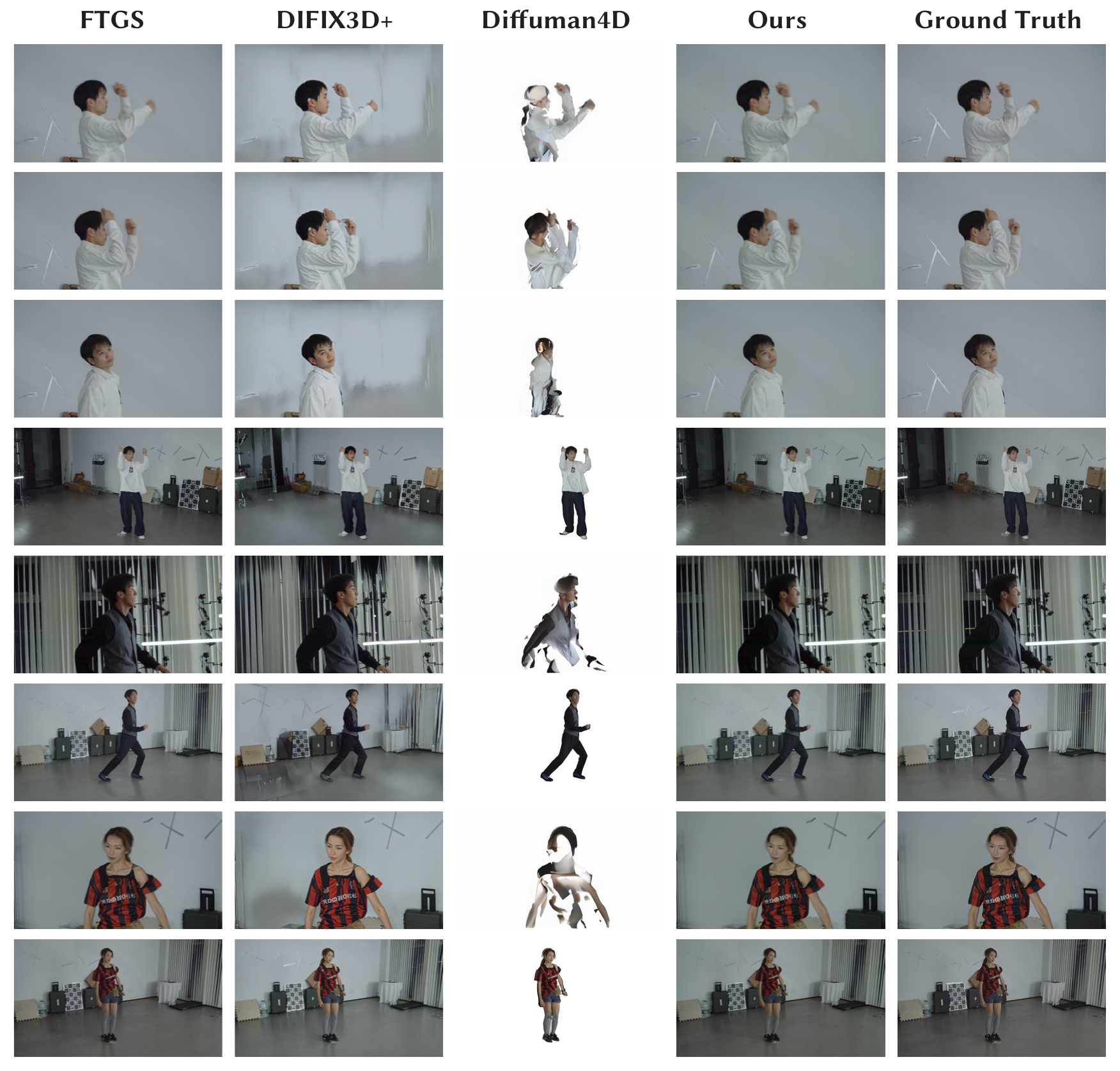}
    \caption{\textbf{Additional qualitative comparisons for novel view synthesis on \textit{RealVV}.}
    We compare FTGS~\cite{Wang2025FreeTimeGSFreeGaussian}, DIFIX3D+~\cite{Wu2025DIFIX3DImproving}, Diffuman4D~\cite{Jin2025Diffuman4D}, RealMe, and the ground truth under the same sparse-capture setting.
    The examples cover both short-focus and long-focus views.
    RealMe better preserves scene backgrounds, human appearance, and view-dependent details.}
    \Description{Additional qualitative comparisons for novel view synthesis on RealVV.}
    \label{fig:supp_nvs_realvv}
\end{figure*}

We provide additional qualitative comparisons on \textit{RealVV} and DNA-Rendering~\cite{Cheng2023DNARendering}.
These examples extend the main paper results to more sequences and viewpoints, covering novel view synthesis, detail enhancement, and 4D-consistent reconstruction under the same limited-capture setting.
As shown in \cref{fig:supp_nvs_realvv}, RealMe preserves both foreground motion and scene background across diverse camera views, while the baselines either lose scene content or introduce visible appearance artifacts.

\begin{figure*}[!t]
    \centering
    \includegraphics[width=\textwidth,height=0.88\textheight,keepaspectratio]{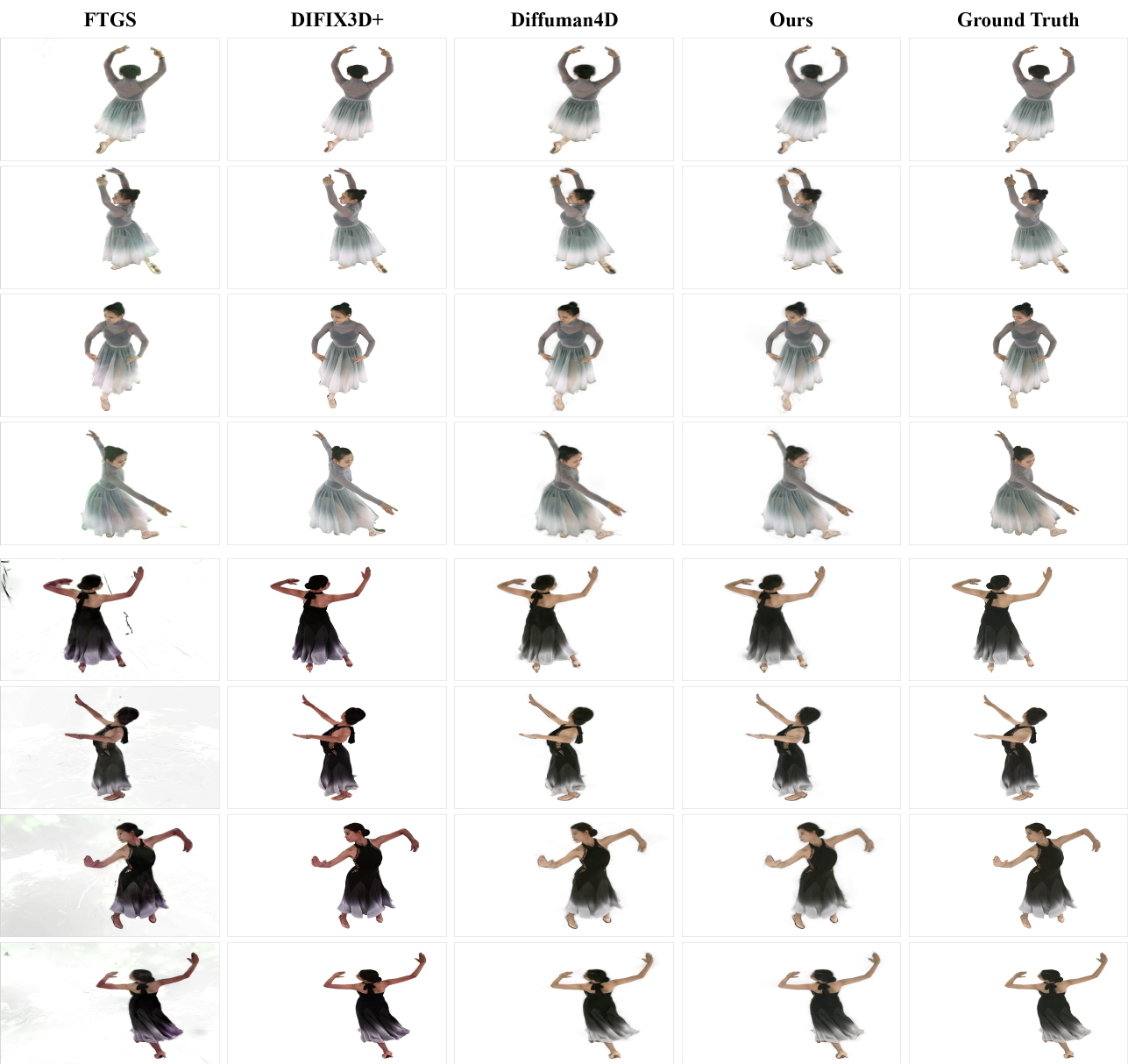}
    \caption{\textbf{Additional qualitative comparisons for novel view synthesis on DNA-Rendering.}
    We compare FTGS~\cite{Wang2025FreeTimeGSFreeGaussian}, DIFIX3D+~\cite{Wu2025DIFIX3DImproving}, Diffuman4D~\cite{Jin2025Diffuman4D}, RealMe, and the ground truth using eight input views.
    The top and bottom four rows show two evaluation sequences across different held-out viewpoints and timestamps.}
    \Description{Additional qualitative comparisons for novel view synthesis on DNA-Rendering.}
    \label{fig:supp_nvs_dna}
\end{figure*}

We also test zero-shot generalization on DNA-Rendering~\cite{Cheng2023DNARendering} in \cref{fig:supp_nvs_dna}.
The examples cover two unseen performers at different timestamps and held-out viewpoints, complementing the quantitative comparison in the main paper.
Across these settings, RealMe preserves each performer's appearance and motion structure while remaining close to the ground truth.

\begin{figure*}[!t]
    \centering
    \includegraphics[width=\textwidth,height=0.88\textheight,keepaspectratio]{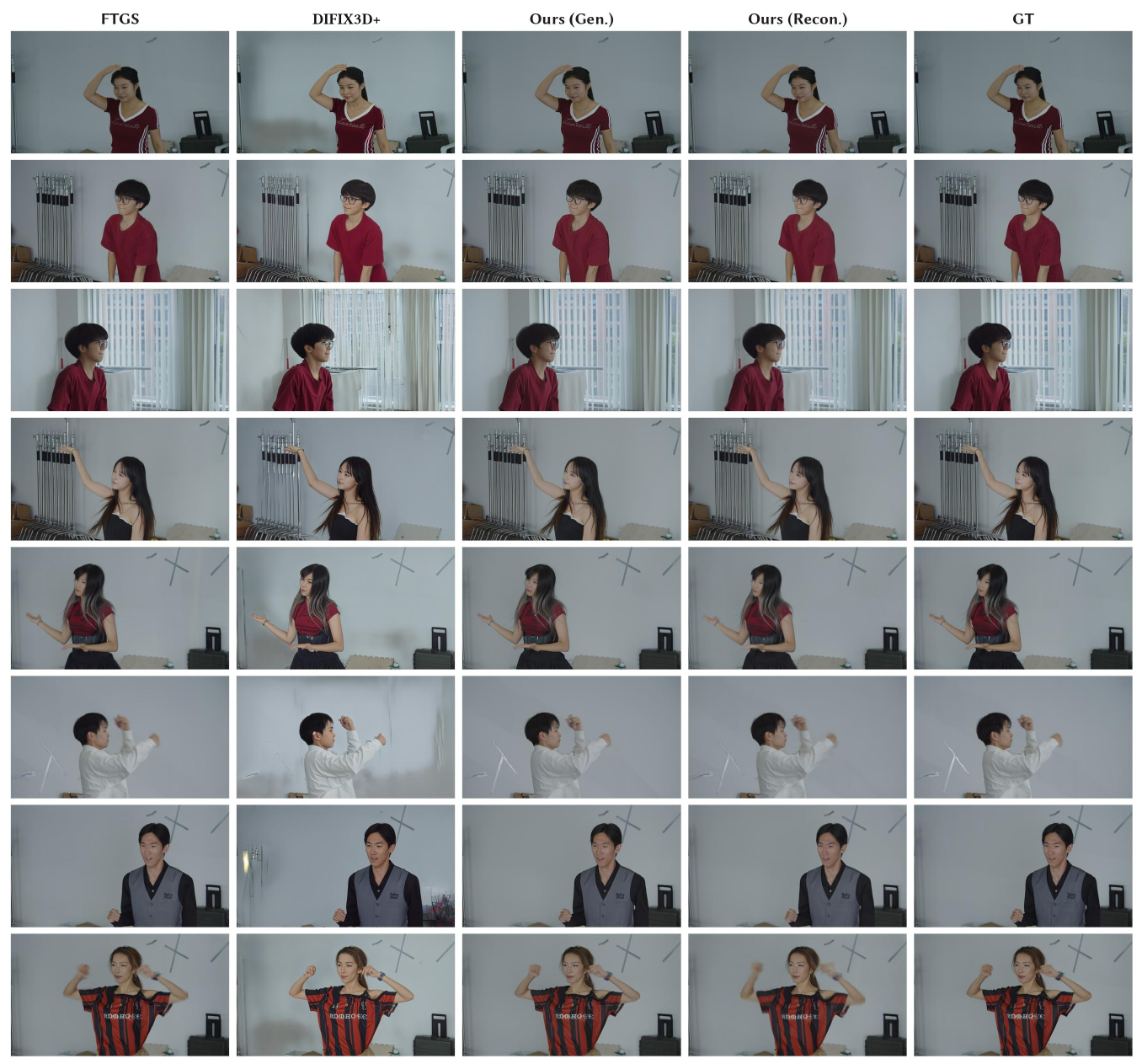}
    \caption{\textbf{Additional qualitative comparisons on long-focus views.}
    We compare FTGS~\cite{Wang2025FreeTimeGSFreeGaussian}, DIFIX3D+~\cite{Wu2025DIFIX3DImproving}, generated views from RealMe, 4D reconstruction from RealMe, and the ground truth.
    Long-focus views stress the ability to recover high-resolution appearance under limited input coverage.
    RealMe produces cleaner details and more faithful scene appearance in these zoomed-in views.}
    \Description{Additional qualitative comparisons on long-focus views.}
    \label{fig:supp_longfocus}
\end{figure*}

We also provide a focused comparison on zoomed-in long-focus views in \cref{fig:supp_longfocus}.
These views are challenging because sparse captures provide weak constraints for fine-scale appearance and background regions.
The comparison shows that the generated views and the reconstructed results from RealMe remain closer to the ground truth than the optimization-based and generation-based baselines.

\subsection{High-Resolution Detail Comparisons}

We include high-resolution detail comparisons on \textit{RealVV}.
Each example presents the full-frame result together with zoomed-in crops, making facial details, clothing textures, and local motion details easier to inspect.
These comparisons complement the quantitative detail enhancement metrics in the main paper.

\subsection{Qualitative Ablation Results}

\begin{figure*}[!t]
    \centering
    \includegraphics[width=\textwidth,height=0.88\textheight,keepaspectratio]{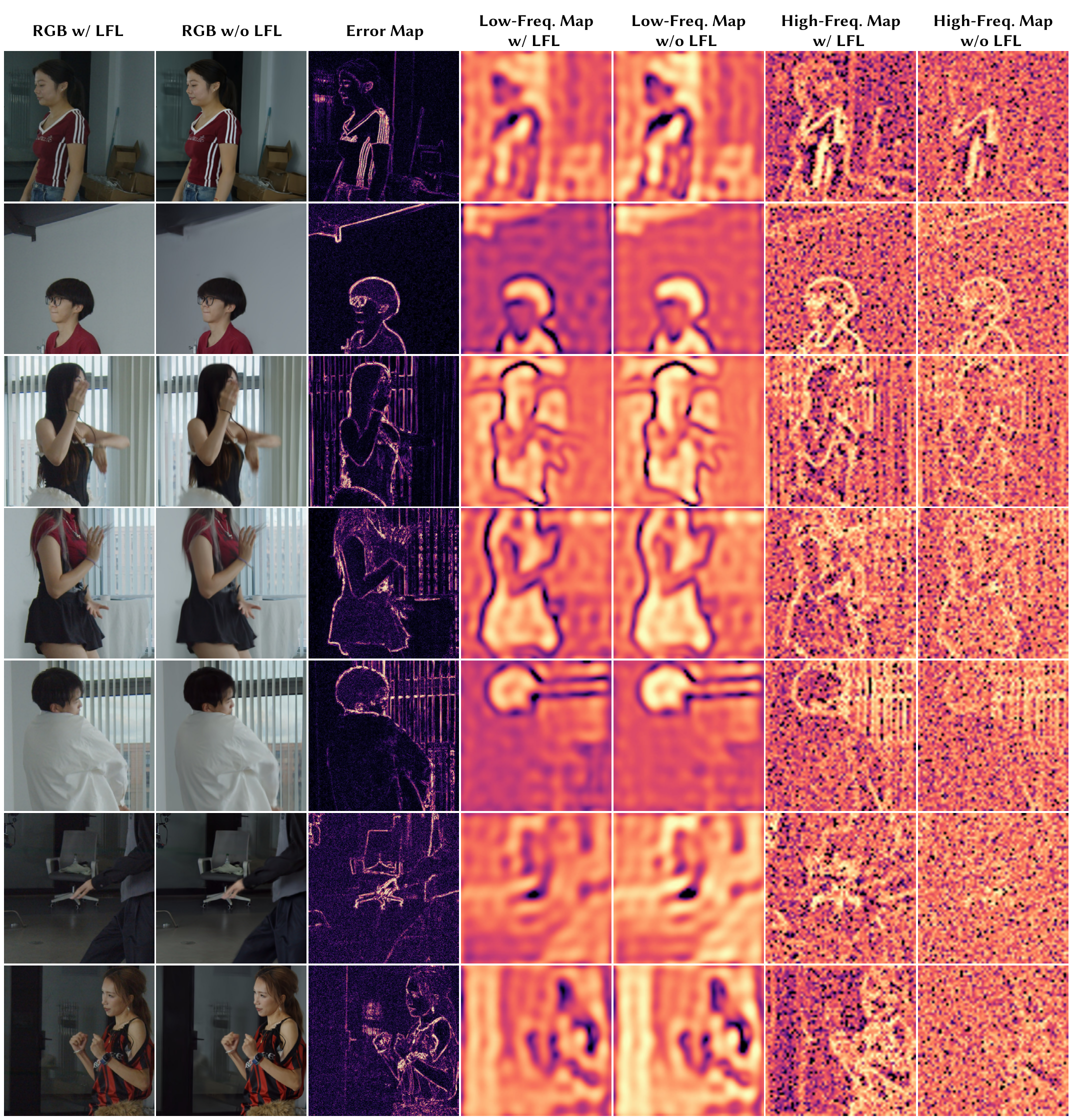}
    \caption{\textbf{Qualitative ablation on latent frequency loss.}
    We compare generated RGB results with and without latent frequency loss (LFL), the RGB error map, and the corresponding low-frequency and high-frequency latent visualizations.
    The latent maps show that LFL better preserves high-frequency structures such as silhouettes, facial details, and clothing boundaries, complementing the quantitative ablation in the main paper.}
    \Description{Qualitative ablation on latent frequency loss.}
    \label{fig:supp_lfl_ablation}
\end{figure*}

We provide qualitative ablation results for layout-aware positional embedding, factorized spatio-temporal attention, and latent frequency loss.
These visual comparisons complement the ablation tables in the main paper by showing how each component affects local detail quality and 4D consistency across time and views.
In \cref{fig:supp_lfl_ablation}, we visualize the effect of latent frequency loss in both RGB space and latent frequency space.
The low-frequency maps capture coarse structure, while the high-frequency maps highlight local texture and boundary details.
With latent frequency loss, the generated results retain sharper high-frequency structures and reduce visible detail errors.

\section{Dataset Visualization}

\begin{figure*}[!t]
    \centering
    \includegraphics[width=\textwidth,height=0.68\textheight,keepaspectratio]{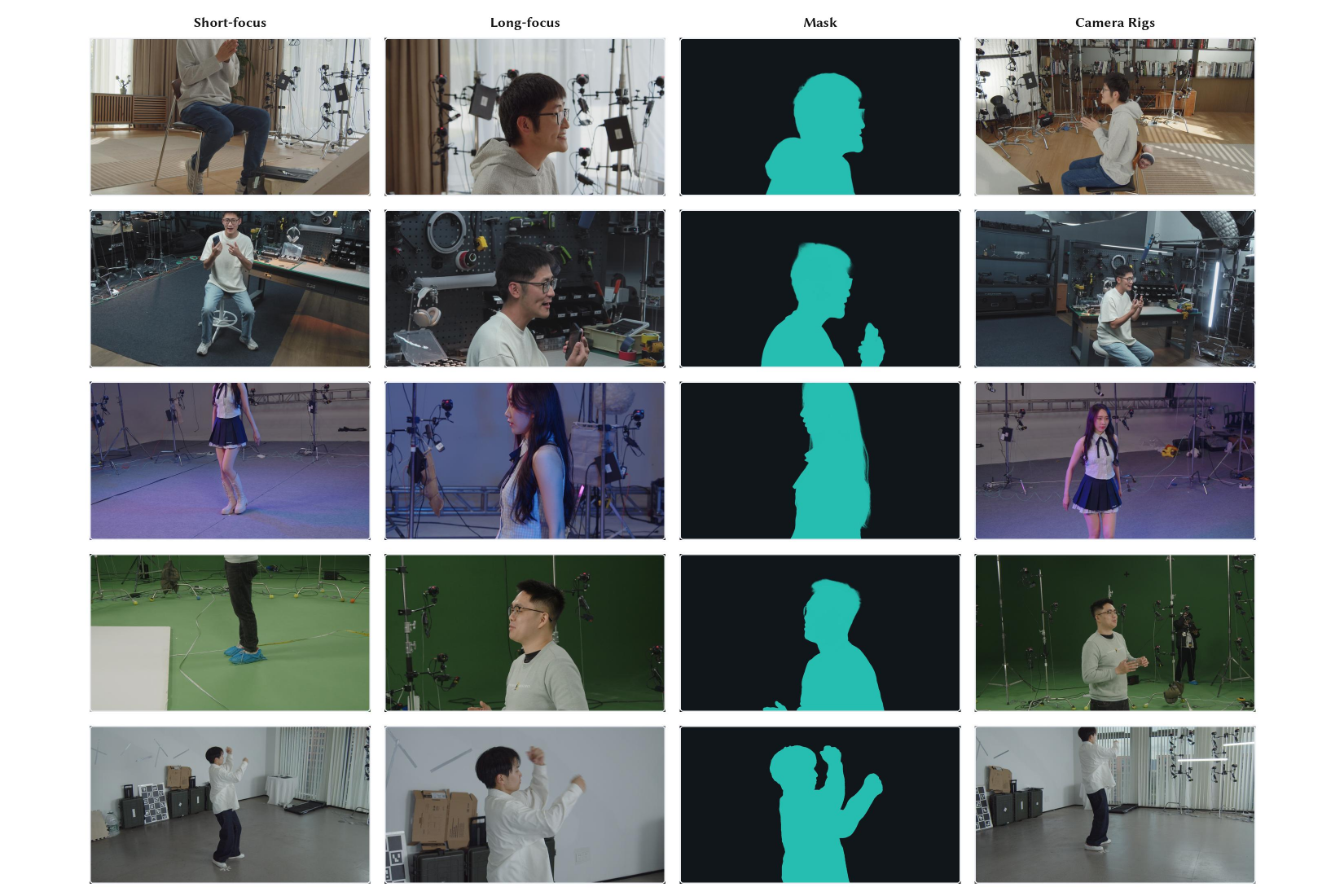}
    \caption{\textbf{RealVV dataset overview.}
    We show representative samples from \textit{RealVV}, including short-focus captures, long-focus captures, foreground masks, and camera-rig views across different scenes.}
    \Description{RealVV dataset overview with representative short-focus captures, long-focus captures, foreground masks, and camera-rig views.}
    \label{fig:supp_realvv_dataset_overview}
\end{figure*}

As shown in \cref{fig:supp_realvv_dataset_overview}, we provide representative captures from \textit{RealVV} to illustrate the dataset diversity.
The overview includes short-focus views that capture the full performer and surrounding scene, long-focus views that preserve local clothing and facial details, foreground masks used for human-centric reconstruction, and camera-rig views that show the multi-view capture setup.
These examples cover different scenes, actors, actions, camera baselines, and focal settings, making the training and evaluation data distribution more concrete.

\end{document}